\documentclass[conference]{IEEEtran}
\IEEEoverridecommandlockouts
\usepackage{cite}
\usepackage{amsmath,amssymb,amsfonts}
\usepackage{algorithmic}
\usepackage{graphicx}
\usepackage{textcomp}
\usepackage{xcolor}
\usepackage{eso-pic}
\def\BibTeX{{\rm B\kern-.05em{\sc i\kern-.025em b}\kern-.08em
    T\kern-.1667em\lower.7ex\hbox{E}\kern-.125emX}}
\begin{document}

\title{NAS-Driven Hardware Accelerator Exploration for Edge AI and Quantization Effects on the Pareto Space\\
\thanks{This work has received funding from European Projects: COP-PILOT Grant agreement ID: 101189819, 6G-XCEL Grant agreement ID: 101139194. Funded by the European Union. Views and opinions expressed are however those of the author(s) only and do not necessarily reflect those of the European Union or the Smart Networks and Services Joint Undertaking or the European Commission. Neither the European Union nor the granting authority can be held responsible for them.}
}

\author{\IEEEauthorblockN{Eleftherios Mylonas}
\IEEEauthorblockA{\textit{Dept. of ECE} \\
\textit{University of Patras}\\
Patras, Greece \\
0000-0002-7676-685X}
\and
\IEEEauthorblockN{Angelos Kouprizas}
\IEEEauthorblockA{\textit{Dept. of ECE} \\
\textit{University of Patras}\\
Patras, Greece \\
0009-0000-6577-7142}
\and
\IEEEauthorblockN{Michael Birbas}
\IEEEauthorblockA{\textit{Dept. of ECE} \\
\textit{University of Patras}\\
Patras, Greece \\
0000-0002-6124-221X}
\and
\IEEEauthorblockN{Alexios Birbas}
\IEEEauthorblockA{\textit{Dept. of ECE} \\
\textit{University of Patras}\\
Patras, Greece \\
0000-0002-9468-7215}
}

\newlength{\origtextheight}
\setlength{\origtextheight}{\textheight}
\addtolength{\textheight}{-0.1in}

\AddToShipoutPictureBG*{%
  \AtPageLowerLeft{%
    \hspace*{\dimexpr(\paperwidth-\textwidth)/2\relax}%
    \raisebox{0.3in}{\parbox[b]{\textwidth}{\fontsize{8}{9.5}\selectfont\noindent © 2026 IEEE.  Personal use of this material is permitted.  Permission from IEEE must be obtained for all other uses, in any current or future media, including reprinting/republishing this material for advertising or promotional purposes, creating new collective works, for resale or redistribution to servers or lists, or reuse of any copyrighted component of this work in other works.}}%
  }%
}

\maketitle

\begin{abstract}
Edge AI deployment demands neural architectures that are simultaneously accurate, computationally efficient, and hardware-deployable — a challenge addressed by hardware-aware Neural Architecture Search (NAS). While recent works incorporate quantization directly into the NAS loop, these approaches expand search complexity and tightly couple architecture and quantization design. The simpler post-search quantization strategy has received little analytical attention: the effects of Post-Training Quantization (PTQ) on the NAS-discovered Pareto structure remain uncharacterised, and no framework combines quantized architecture mapping onto reconfigurable accelerators with automated hardware exploration. This paper addresses both gaps. First, a three-stage pipeline is proposed: a hardware-agnostic Pareto rank surrogate frontend on NAS-Bench-201, a quantization bridge with Pareto-aware filtering and feedback control, and an evolutionary Domain Space Exploration (DSE) backend on CGRA4ML for optimal hardware mapping. Second, an empirical study characterises how INT4 PTQ perturbs the NAS-Bench-201 Pareto space through formal stability metrics on ground-truth data for all 15,625 architectures, and demonstrates that an FP32 zero-shot surrogate outperforms a dedicated INT4-trained surrogate in Pareto space coverage across two standard search strategies.
\end{abstract}


\section{Introduction}
Artificial Intelligence (AI) deployment has fundamentally changed as a result of deep learning's explosive growth. The emphasis shifted from centralized cloud AI to the more resource-constrained edge AI for more data privacy, inference latency decrement, and autonomous operation in settings where cloud access is not possible. This transition came with lots of computing, memory and power demands towards edge devices, highlighting the need for new types of AI-capable systems, but also new efficient methods of AI task deployment across heterogeneous platforms. The challenge of efficiently designing models that are simultaneously accurate and deployable on diverse hardware — without manual trial-and-error across a combinatorial space of architectural choices — made automated model design an essential research priority, giving rise to the domain of Neural Architecture Search (NAS) \cite{b1}.

NAS automates the AI model design, testing, search and selection process in various ways, either by testing different neural network layer combinations from the start or based on benchmarks from pre-trained combinations. However, it is still an open challenge, especially when it comes to edge AI deployment. Benchmark coverage remains limited: most NAS benchmarks target a small set of tasks and focus on full-precision (FP32) model accuracies, completely neglecting other hardware performance metrics. Exceptions like HW-NAS-Bench \cite{b2} capture performance data from a fixed set of hardware targets which operate almost exclusively with FP32 precision, and do not examine the effects of quantization — which is the standard practice during any AI model deployment on hardware platforms — on NAS-discovered architectures. Search efficiency is also non-trivial: naive exhaustive search over large spaces is intractable, and without a principled estimation strategy, NAS may converge to suboptimal architectures. Performance estimation is also dependent on how an architecture is encoded for the NAS process, since a proper encoding reveals the most useful model structural and operational data. Lastly, no existing work has examined the effect of quantization on multi-objective performance estimators or offers a hardware Design Space Exploration (DSE) framework targeting flexible reconfigurable accelerator architectures for edge deployment.

This paper addresses these challenges via a three-stage hardware-aware NAS workflow that identifies optimal candidate architectures and maps them onto a configurable Coarse Grain Reconfigurable Array (CGRA)-based accelerator. The main contributions of this work are:
\begin{itemize}
\item a three-stage hardware-aware NAS procedure comprising a hardware-agnostic frontend trained on NAS-Bench-201 \cite{b3} and CIFAR-10 \cite{b4} that identifies the best candidate architectures in full precision, the quantization bridge that applies the quantization scheme and filters out the failing architectures, and a hardware-aware backend that searches for the optimal CGRA architecture per surviving architecture via evolutionary optimisation;
\item an empirical study of the effect of INT4 Post Training Quantization (PTQ) via the open-source Brevitas on NAS-Bench-201 architectures, including formal Pareto stability metrics computed on ground-truth quantized data for the full 15,625-architecture space, a quantization-aware surrogate providing a three-way transferability analysis of FP32 Pareto rank predictions under quantization-induced distribution shift, and a search comparison demonstrating that the FP32 zero-shot surrogate outperforms a dedicated INT4-trained surrogate in normalised hypervolume across two different strategies;
\item a hardware DSE environment built on the open-source CGRA4ML \cite{b5} that finds the optimal hardware mapping for each quantized architecture surviving the quantized Pareto filter, using a free analytical performance oracle and a three-term normalised scalar fitness function.
\end{itemize}

The rest of the paper is structured as follows: Section II further strengthens the theoretical background and reviews the current state-of-the-art; Section III presents the proposed NAS system and explains each stage analytically; Section IV reports the results; Section V concludes and outlines future work.

\setlength{\textheight}{\origtextheight}
\section{Theoretical Background}

\subsection{NAS}
\begin{figure}[t]
\centerline{\includegraphics[width=19pc]{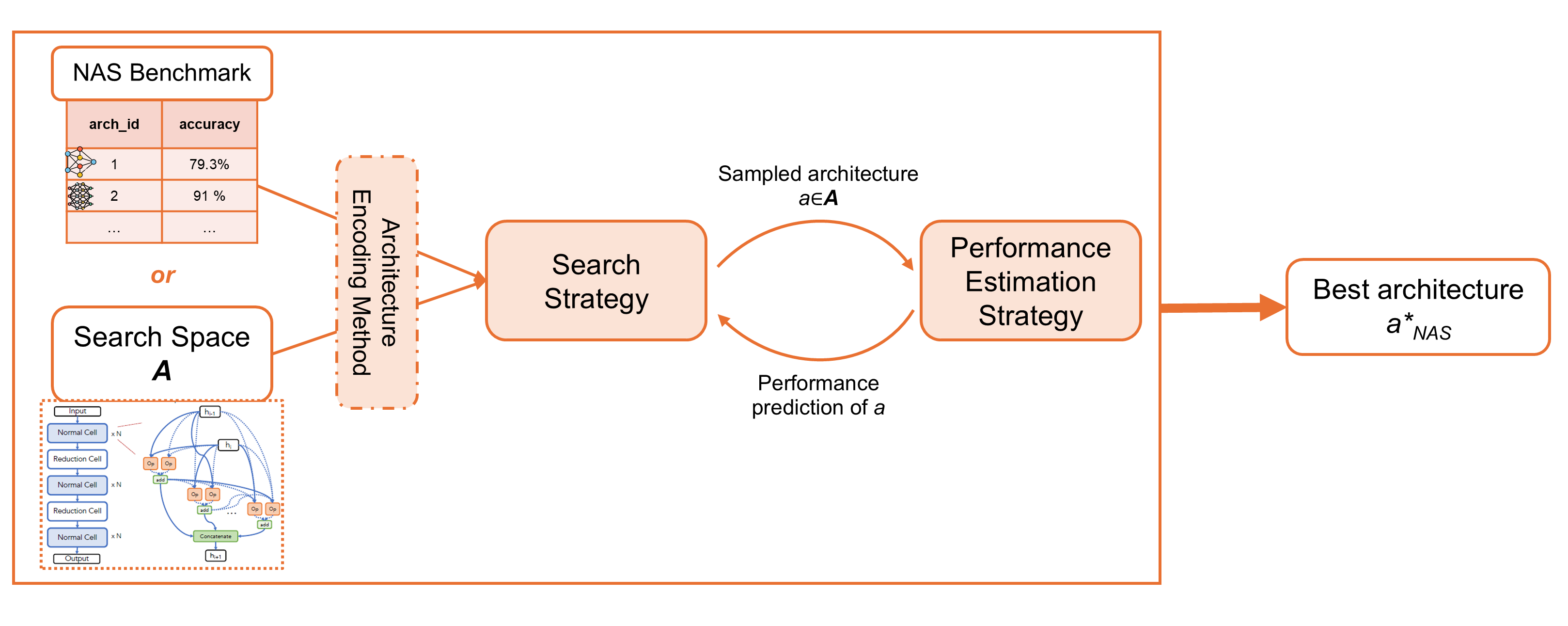}}
\caption{Overview of a general NAS process.}
\label{fig1}
\end{figure}

NAS is a subfield of Automated Machine Learning (AutoML) that aims to automate the design of neural network architectures. A typical NAS procedure as shown in Fig. \ref{fig1} consists of three components: a \textit{search space} defining the set of candidate architectures (e.g., cell-based spaces with a fixed set of operations and connectivity patterns); a \textit{search strategy} that samples candidate architectures from that space (evolutionary algorithms, reinforcement learning, etc.); and a \textit{performance estimation strategy} that evaluates or predicts each candidate's quality according to one or more metrics, most commonly its accuracy. In addition to this, an \textit{architecture encoding method} is also a common practice for transforming a candidate architecture into a vector that encompasses all its structural details and can lead to more accurate performance prediction. Typical NAS can be mathematically expressed as an optimization problem as follows:

\begin{equation}
a^{*}_{NAS},w^{*}_{NAS}=\arg\max_{a,w}ACC_{val}(a,w),\label{eq1}
\end{equation}

where $a,w$ are one candidate architecture and its weight parameters respectively and $a^{*}_{NAS},w^{*}_{NAS}$ are the winning architecture and weights respectively for which the validation accuracy is maximized across the search space.

To enable reproducible comparison and avoid the prohibitive cost of training each candidate from scratch, the community has developed \textit{NAS benchmarks} — collections of pre-trained architectures from a specific search space with associated metrics — with some of the most widely used being NAS-Bench-201 \cite{b3} and NAS-Bench-360 \cite{b7}. These benchmarks function as lookup tables that allow any NAS method to evaluate candidate architectures by querying pre-computed results, thus enabling performance comparison between different NAS methods, search strategies, etc.

\subsection{Hardware-aware NAS}
Hardware-aware NAS can be seen as the multi-objective version of the optimization problem described in \eqref{eq1} for a set of hardware performance metrics. If we assume that $f_i$ are the performance metrics with $i\in[1,...,n]$ being the metric index and $n$ the total number of metrics, the problem can be formulated as:

\begin{equation}
a^{*}_{NAS},w^{*}_{NAS}=\arg\max_{a,w}f_{1}(a, w),...,f_{n}(a, w)\label{eq2}
\end{equation}

When working on top of a NAS benchmark, where $w$ is fixed per candidate architecture $a$, the problem is simplified into:

\begin{equation}
a^{*}_{NAS}=\arg\max_{a}f_{1}(a),...,f_{n}(a)\label{eq3}
\end{equation}

Equation \eqref{eq3} can be turned into a constrained optimization problem, taking into account the user-assigned computation budget and their desired performance metrics.

The most prominent works in this field \cite{b8},\cite{b9} search for a set of solutions for which no further optimization is possible, that is, these solutions cannot be dominated by others. This non-dominated set is called the \textit{Pareto front}. For an architecture $a$ to dominate another architecture $b$ $(a>b)$, the following should be true:

\begin{equation}
\forall i \in [1,...,n], f_{i}(a) \leq f_{i}(b) \wedge \exists j f_{j}(a) < f_{j}(b)\label{eq4}
\end{equation}

NAS search strategies that try to find Pareto front architectures use special performance estimators called Pareto rank predictors or Pareto rank surrogates. A great work on Pareto rank surrogates and appropriate architecture encoding is done in \cite{b9}, to which we based our work on this paper.

\subsection{State-of-the-art \& Related work}
Hardware-aware NAS has been significantly advanced by the introduction of dedicated benchmarks. HW-NAS-Bench extends NAS-Bench-201 with latency and energy measurements across six hardware platforms, enabling hardware-aware architecture comparison without physical hardware access during search. However, all metrics are collected in FP32 — no quantization is applied, and its effect on Pareto rankings is not studied. On the surrogate side, HW-PR-NAS \cite{b9} addresses the dominance errors produced by independent per-objective surrogates in multi-objective NAS, proposing instead a surrogate trained directly on Pareto rank labels with a listwise ranking loss, which preserves the multi-objective ordering of architectures. This approach forms the foundation for the Stage I surrogate in this paper.

Recognising that FP32 NAS results may not transfer well to quantized deployment, a substantial body of work incorporates quantization directly into the NAS search loop. HAQ \cite{b10} uses reinforcement learning to assign mixed-precision bit-widths per layer, APQ \cite{b11} jointly searches architecture and quantization policy, and SimQ-NAS \cite{b12} simultaneously searches sub-network architecture and quantization policy across CNN and transformer architectures. More recently, QuantNAS \cite{b13} searches directly from a fully quantized supernet to address the suboptimality of post-search quantization. These methods treat quantization as part of the search problem itself, substantially expanding the search space. In parallel, hardware accelerator co-search has also been explored: NAAS \cite{b14} proposes a two-level co-search over neural architectures and accelerator configurations, demonstrating that joint search yields better accuracy-efficiency trade-offs than sequential approaches.

Despite this progress, a critical gap remains. Joint quantization-NAS methods avoid the problem of post-search quantization by incorporating bit-width into the search; hardware-aware NAS benchmarks ignore quantization entirely; deployment-focused works apply PTQ without studying its effect on relative architectural rankings. No existing work characterises how PTQ perturbs the Pareto structure of FP32 NAS-discovered architectures, nor provides formal stability metrics for this perturbation at full search-space scale. This paper addresses this gap through two distinct contributions: a quantization bridge that integrates INT4 PTQ and budget-aware filtering with a feedback loop into the NAS pipeline, and a systematic empirical study — the first of its kind — that characterises how INT4 PTQ perturbs the Pareto structure of NAS-Bench-201 architectures through formal stability metrics and a three-way surrogate transferability analysis.

\section{System Overview}
\begin{figure}[b]
\centerline{\includegraphics[width=19pc]{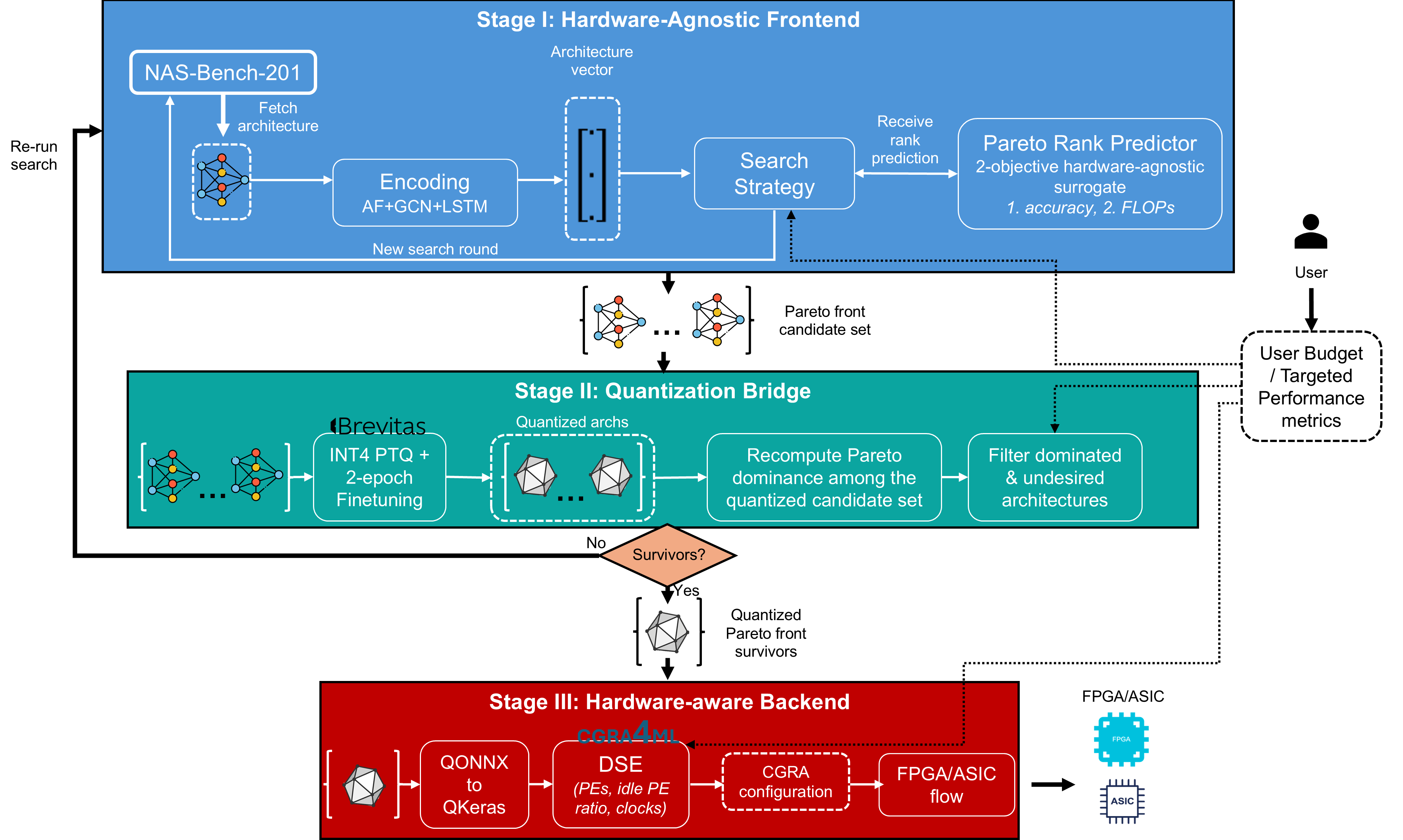}}
\caption{Full hardware-aware NAS pipeline.}
\label{fig2}
\end{figure}

The complete NAS system proposed in this work is presented in Fig. \ref{fig2}. It consists of three stages and is considered a NAS-then-quantized approach. This approach is chosen since it is relatively straightforward for deployment, it decouples the model and hardware accelerator optimization problem from each other, and it enables separate frontend and backend solutions for NAS and hardware design and exploration respectively. 

\subsection{Stage I: Hardware-Agnostic Frontend}
The first stage of the system is the Hardware-Agnostic Frontend and is responsible for the main NAS procedure. It is based on a redesign of HW-PR-NAS \cite{b9}, where the architecture encoding module was trained only for the NAS-Bench-201 search space and its LSTM submodule was enhanced with a Pool layer as the output node. This was done to increase Kendall's Tau (KT) correlation results between the predicted scores of the Pareto rank predictor and the actual rankings of the architectures. The predictor itself — also called as surrogate — was trained so as to work with two hardware-agnostic objectives, accuracy and FLOPs, where accuracy is directly derived by the NAS-Bench-201 while the FLOPs are directly extracted from the models themselves. The NAS search strategies tested are the Random Search (RS) and the Multi-Objective Evolutionary Algorithm (MOEA). When running both searches, a target value for FLOPs is inserted so as to guide them towards the right region of the Pareto space. After one run, this stage returns ten architectures which have been evaluated as the Pareto front candidate set.

\subsection{Stage II: Quantization Bridge}
The second stage of the system is the Quantization Bridge, and can be assumed as a middleware between the hardware-agnostic and the hardware-aware part of the system. It is responsible for the quantization of the candidate architectures, their Pareto re-ranking and the final filtering of dominated and undesired architectures that can arise after the Pareto space shift during quantization. The quantization task is run with the well-established open-source Brevitas tool from AMD-Xilinx. In addition, there is a feedback loop to Stage I in case no survivors are left after this stage, for which a new search is triggered. For the quantization part, the INT4 quantization strategy was chosen. The reasoning is presented in the respective subsection of Section \ref{results}. 

\subsection{Stage III: Hardware-Aware Backend}
The third stage of the system is the Hardware-Aware Backend and it is responsible for the optimal mapping of the survivor models to customized hardware accelerator architectures. It is based around the open-source CGRA4ML tool, which is a set of Python and SystemVerilog scripts able to compile QKeras models into deployment-ready, flexible CGRA designs. In order to work on the quantized models from Brevitas into CGRA4ML, we implement a QONNX-to-QKeras model translation layer \cite{b15} which translates the structure of the model into a QKeras CGRA4ML-compatible format and appropriately copies the quantized weight parameters into this new model. CGRA4ML supports different mixed-precision integer quantization modes (INT1/2/4/8 for activations/weights, INT8/16/32 for biases) and is configurable in many ways, from the number of Processing Elements (PEs) to the total internal memory bank and the input/output AXI port widths, thus it is an ideal choice. In order to test different designs for each survivor model and decide on the best architecture, a DSE procedure for CGRA4ML is prepared. In order to test different designs for each survivor model and decide on the best architecture, a DSE procedure for CGRA4ML is prepared. The DSE explores four configuration parameters: PE array rows $R \in \{8, 10, 12, 16\}$, columns $C \in [12, 96]$ in steps of the dominant kernel width, weight SRAM depth $d_w \in [256, 1024]$ in steps of 64, and AXI width $w_{AXI} \in \{64, 128\}$, subject to a hard resource constraint $R \times C \leq MAX\_PEs$ that reduces the valid space to $\approx$1,200 combinations.

An evolutionary algorithm evaluates each candidate configuration using the free CGRA4ML analytical oracle \textit{predict\_model\_performance()}, which returns inference clock cycles, PE utilisation, and idle PE ratio without triggering RTL generation or synthesis. A three-term normalised scalar fitness jointly minimises latency, PE idle ratio, and array area:

\begin{equation}
\begin{aligned}
   \;& F = -\frac{clocks}{clocks_{ref}} - \alpha(1-\overline{util}) - \beta\frac{R \times C}{MAX\_PEs}\\
   & - \lambda \cdot \max(0, R{\times}C - MAX\_PEs) \times 10^{4} 
\end{aligned}
\label{eq5}
\end{equation}

with $\alpha{=}0.2$, $\beta{=}0.1$, and $\lambda$ as the hard constraint penalty. All terms are normalised to $\approx[0,1]$ before weighting. The algorithm uses tournament selection, constraint-preserving mutation, and elite carryover, terminating when fitness improvement falls below $10^{-4}$ over 8 generations. Finally, the winning CGRA configuration is exported so as to be tested on FPGA or ASIC flows.

\section{Experimental Results} \label{results}
Having explained the proposed three-stage hardware-aware NAS system, the next step is to present the results obtained from its evaluation.

\subsection{Quantization Scheme Selection}

\begin{figure}[t]
\centerline{\includegraphics[width=19pc]{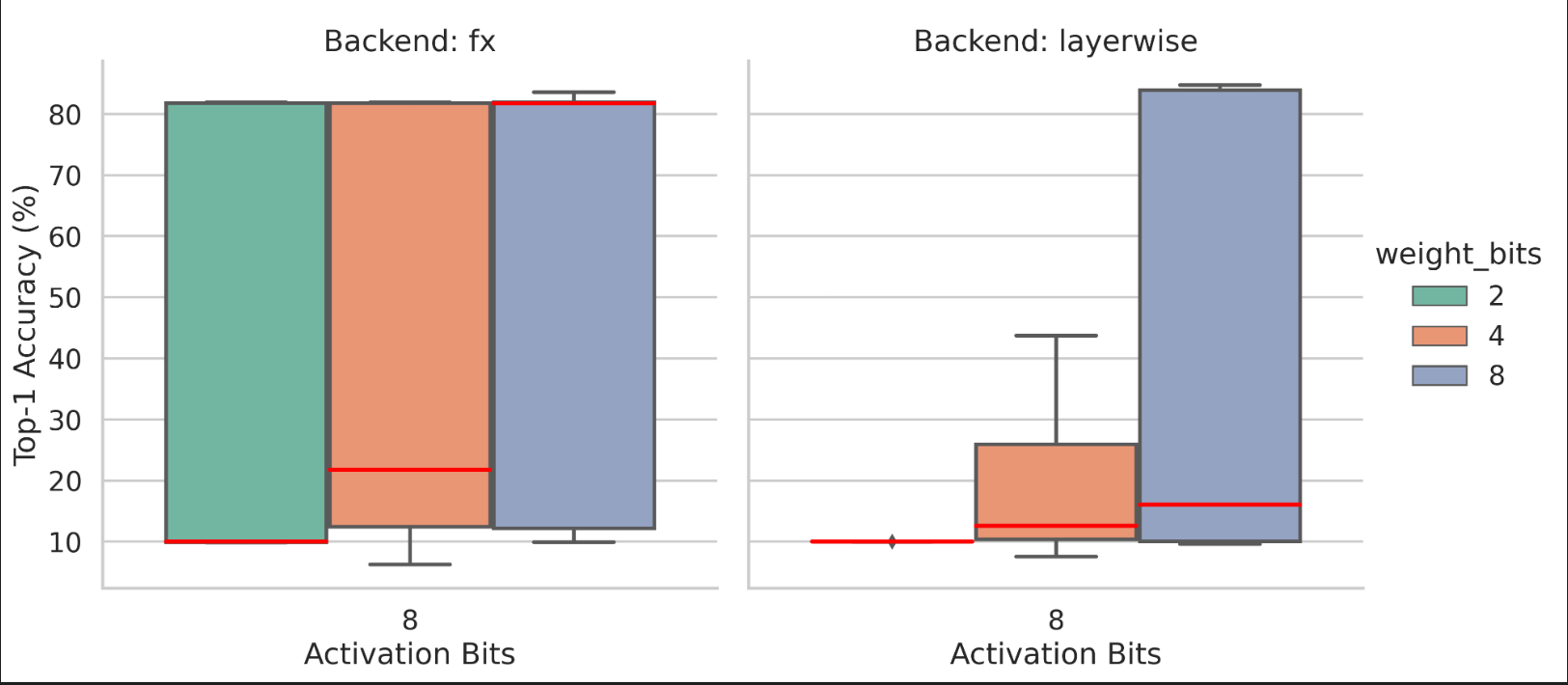}}
\caption{Statistical Analysis of INT8 PTQ.}
\label{fig3}
\end{figure}

In order to choose a quantization strategy, a sample architecture was used as a baseline with its size and complexity as the selection criteria. This was done so as to ensure that a set of complexity-robust PTQ methods is identified. A total of 325 PTQ configurations were tested without finetuning, of which six configurations were selected as the most promising, summarized in Table \ref{tab1}. It is necessary to mention that the six Brevitas PTQ features shown in the table columns were the most prominent ones with equally important contribution to the PTQ results. Most of the identified quantization schemes use 8-bit activation quantization, which shows the superiority of INT8 PTQ. This is further supported by statistical analysis of different INT8 PTQ runs as shown in Fig. \ref{fig3}. However, INT4 PTQ showed also great versatility with half of INT8 PTQ's complexity, with further runs showing that a simple two-epoch finetuning achieves accuracy restoration from 9.49\% to even 83.53\%, proving that the original model information is preserved. For this reason, PTQ Id 6 was chosen as the quantization scheme for this work.

\begin{table}[t]
\caption{Complexity-robust PTQ Configurations$^{\mathrm{*}}$}
\begin{center}
\begin{tabular}{|c|c|c|c|c|c|c|c|}
\hline
\textbf{Id} & \textbf{Act.} & \textbf{Weight} & \textbf{Bias} & \textbf{Back-} & \textbf{Equali-} & \textbf{Act.} & \textbf{Top1}\\
& \textbf{Bits} & \textbf{Bits} & \textbf{Bits} & \textbf{end} & \textbf{zation} & \textbf{Quant.} & \textbf{Acc.}\\
& &  & & & & \textbf{Type} & \textbf{(\%)}\\
\hline
1 & 8 & 8 & None & layer- & fx & asym & 84.76 \\
& &  & & wise & & & \\
\hline
2 & 8 & 2 & 16 & fx & layer- & sym & 81.92 \\
& &  & & & wise & & \\
\hline
3 & 8 & 4 & 16 & fx & layer- & sym & 81.92 \\
& &  & & & wise & & \\
\hline
4 & 8 & 8 & 16 & fx & layer- & sym & 81.92 \\
& &  & & & wise & & \\
\hline
5 & 4 & 8 & None & layer- & layer- & asym & 41.29 \\
& &  & & wise & wise & & \\
\hline
6 & 4 & 4 & None & layer- & layer- & asym & 26.22 \\
& &  & & wise & wise & & \\
\hline
\multicolumn{7}{l}{$^{\mathrm{*}}$Tested on NAS-Bench-201 architecture with id \textbf{15624}.}
\end{tabular}
\label{tab1}
\end{center}
\end{table}

\subsection{Quantization Error Against Model Complexity} \label{quant_flops}

\begin{figure}[t]
\centerline{\includegraphics[width=19pc]{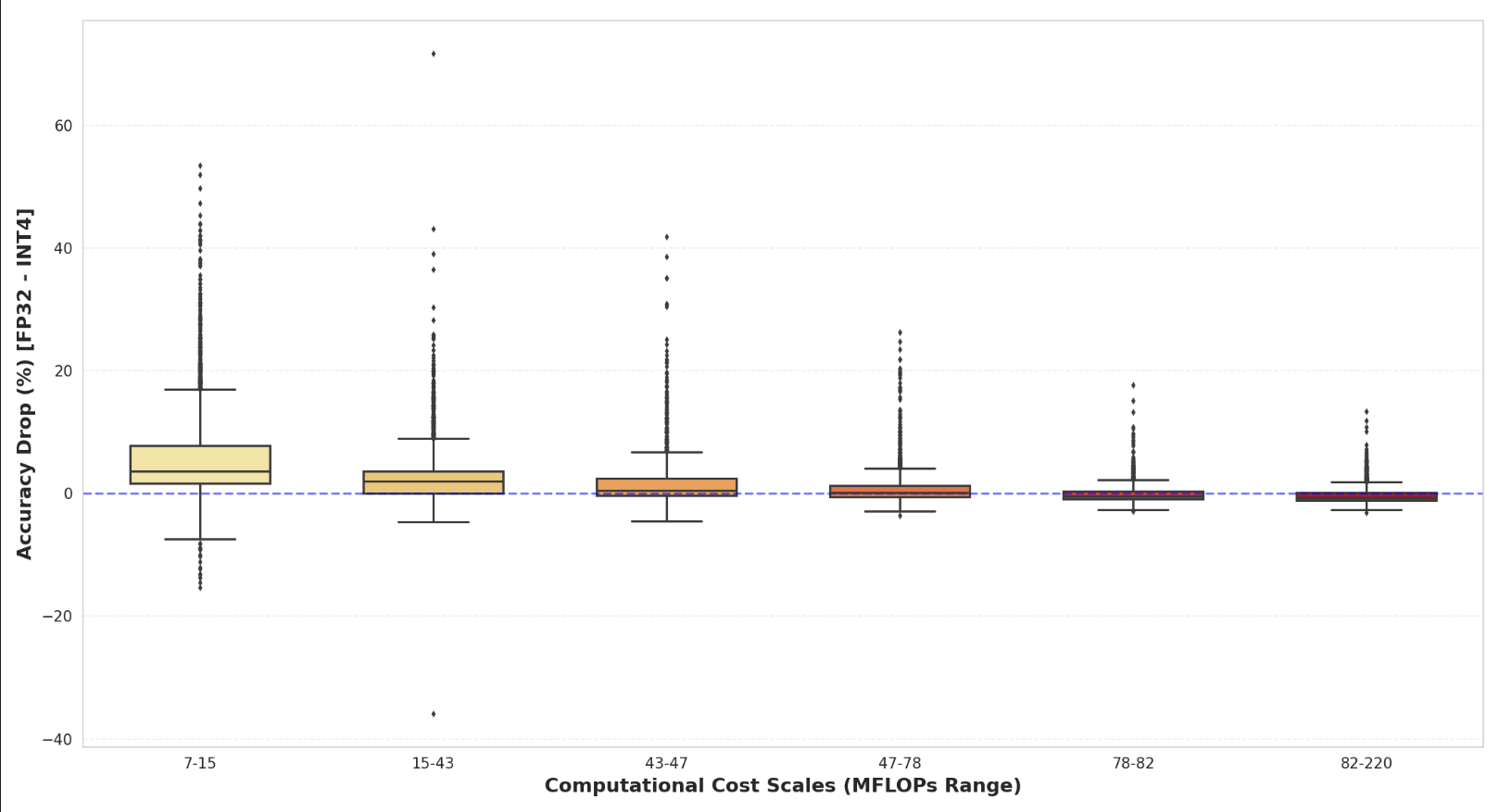}}
\caption{Accuracy error after quantization per MFLOPs.}
\label{fig4}
\end{figure}

\begin{figure}[t]
\centerline{\includegraphics[width=19pc]{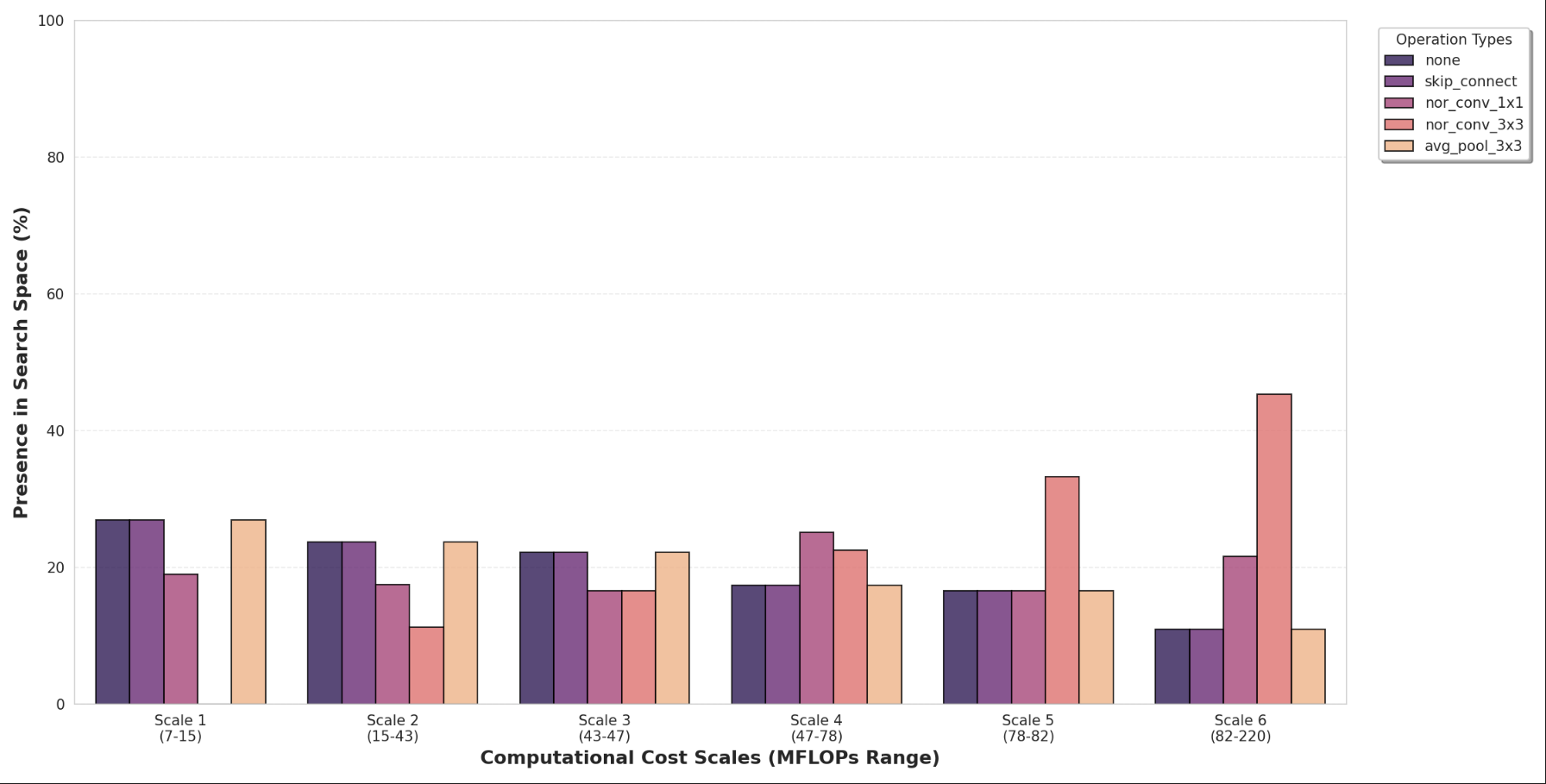}}
\caption{NAS-Bench-201 operations distribution per MFLOPs.}
\label{fig5}
\end{figure}

An interesting observation arises when plotting the accuracy error during quantization against the original complexity of the models in FLOPs, as shown in Fig. \ref{fig4}. Models at lower computational scales (7–47 MFLOPs) exhibit considerably higher sensitivity to INT4 quantization, with median accuracy drops reaching approximately 5\% at the 7–15 MFLOPs range, wide interquartile ranges, and numerous outliers exceeding 50\% degradation. As complexity increases, this sensitivity diminishes progressively, with the median converging to approximately 0\% and variance collapsing substantially. This behavior can be partly attributed to the architectural composition at each scale, as revealed by Fig. \ref{fig5}. At lower MFLOPs ranges, the five NAS-Bench-201 primitive operations are distributed nearly uniformly, producing structurally sparse and heterogeneous cells. The \textit{none} operation eliminates edges entirely, amplifying quantization errors on the remaining paths, while \textit{skip\_connect} propagates noise forward unfiltered. Convolutional operations, by contrast, exhibit inherent lowpass filtering behavior that naturally attenuates the high-frequency noise introduced by reduced-precision arithmetic, an effect most pronounced in \textit{nor\_conv\_3x3} due to its 3×3 spatial receptive field. As scale increases toward 82–220 MFLOPs, \textit{nor\_conv\_3x3} comes to dominate the composition, accounting for nearly 45\% of all operations, while all others recede below 25\%. This structural convergence toward a single quantization-robust operation type directly explains the simultaneous stabilization of accuracy drop observed in Fig. \ref{fig4}, establishing a clear link between architectural homogeneity and robustness to INT4 quantization.

\subsection{Quantization's Impact on Pareto Front \& Surrogate Performance}
\begin{figure}[t]
\centerline{\includegraphics[width=19pc]{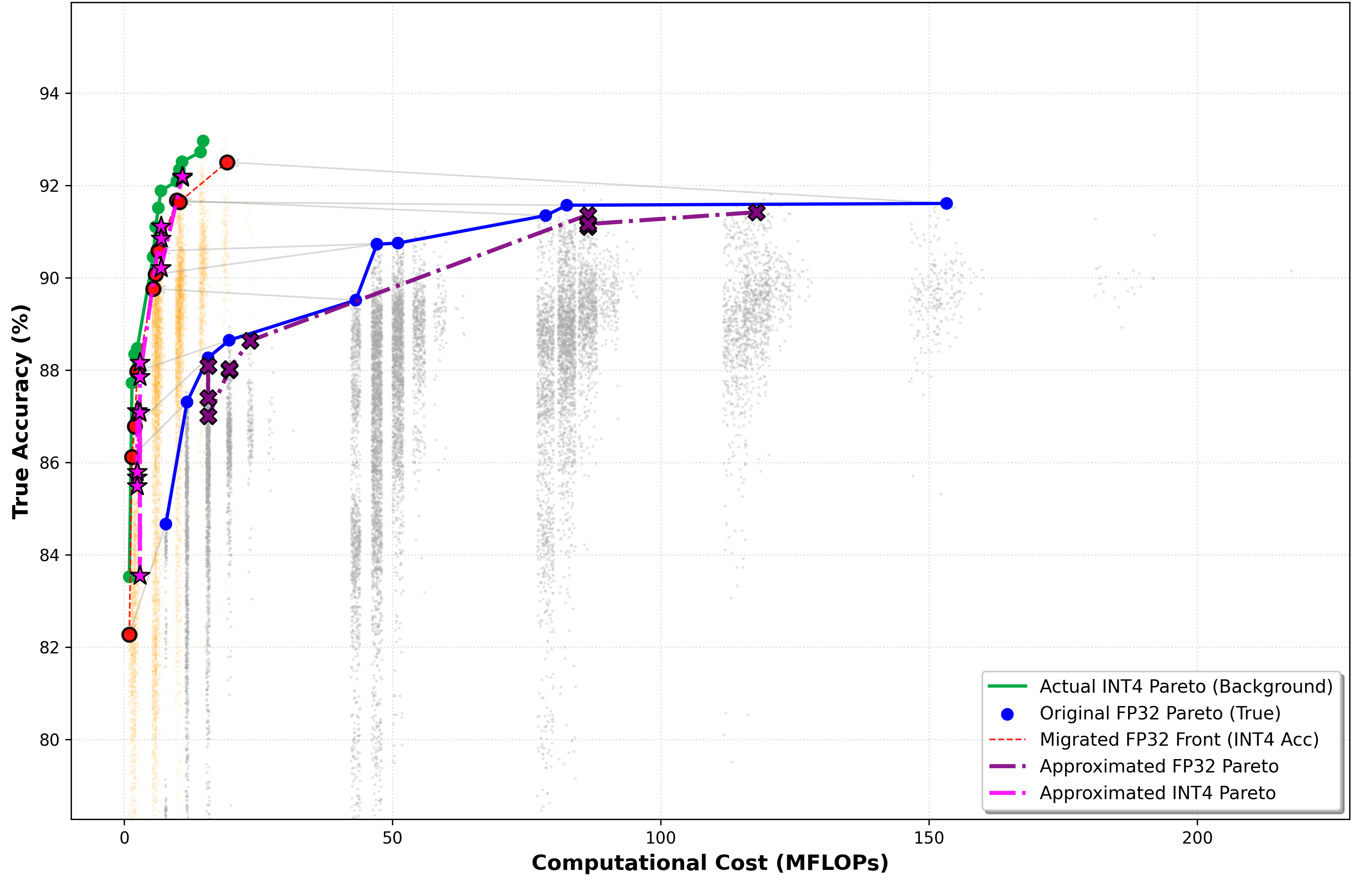}}
\caption{Quantization impact on NAS-Bench-201 Pareto space.}
\label{fig6}
\end{figure}

Fig. \ref{fig6} presents the Pareto front analysis across the NAS-Bench-201 search space, comparing model accuracy against computational cost under both FP32 and INT4 precision regimes. The true FP32 Pareto front establishes the accuracy-efficiency upper bound under full precision, while the green curve defines the corresponding ground truth INT4 Pareto front. The approximated FP32 and INT4 Pareto fronts, shown in dark purple and magenta respectively, closely track their ground truth counterparts across the full computational range, demonstrating that the surrogate reliably predicts quantization-aware performance without exhaustive evaluation. Notably, the red dashed curve — representing the migrated FP32 Pareto front, obtained by re-evaluating the original FP32 Pareto-optimal architectures under INT4 quantization — aligns closely with the ground truth INT4 Pareto front. This visual evidence suggests that the FP32 Pareto structure is largely preserved under quantization, raising the possibility that zero-shot transfer of the FP32 surrogate may be sufficient for effective INT4-aware search.

This observation is further strengthened by the surrogate transferability analysis in Table \ref{tab2}. The surrogate achieves a KT rank correlation of 0.8352 on the original FP32 domain, confirming strong ranking consistency. When transferred zero-shot to the INT4 domain — without any retraining — performance drops only moderately to 0.7219, indicating that the FP32-trained surrogate retains meaningful predictive structure across precision regimes. Upon fine-tuning on quantized fully-trained data, the KT recovers to 0.8219, approaching the original domain performance. The narrow gap between zero-shot and fine-tuned transfer, combined with the visual alignment observed in Fig. \ref{fig6}, suggests that the ranking landscape is sufficiently preserved under INT4 quantization to make zero-shot transfer a viable and practical alternative to full surrogate retraining.

\begin{table}[t]
\caption{Surrogate Transferability Metrics}
\begin{center}
\begin{tabular}{|c|c|}
\hline
KT (original) & 0.8352 \\
\hline
KT (quantized, fully-trained) & 0.8219 \\
\hline
KT (zero-shot) & 0.7219 \\
\hline
\end{tabular}
\label{tab2}
\end{center}
\end{table}

To validate this hypothesis, we conducted a search comparison establishing a MFLOPs budget of 0.625--10, contrasting RS (300 samples) against MOEA (250 queries, population of 150, tournament size of 15, mutation rate of 0.9), averaged over 50 independent runs, using the FP32 zero-shot surrogate against a fully trained INT4-specific surrogate. As reported in Table \ref{tab3}, the FP32 zero-shot surrogate yields higher normalized global hypervolume than the INT4-trained surrogate in both search strategies, with hypervolume ratios of 1.12 and 1.07 for RS and MOEA respectively, corresponding to relative improvements of 12.26\% and 6.77\%. This result confirms and extends the findings of the previous two paragraphs: the FP32 surrogate benefits from a cleaner, less noisy training signal, and because the FP32 and INT4 Pareto structures are substantially correlated, this reliability transfers effectively. The INT4 surrogate seems to prefer higher-FLOPs architectures at the cost of underrepresenting the low-FLOPs region that the hypervolume metric rewards. Accuracy results are nearly identical across both surrogates and search strategies, with all differences falling within one standard deviation, confirming that both approaches find architectures that perform well under both precision regimes simultaneously. 

The stability metrics in Table \ref{tab3} further characterize the broader impact of quantization on the original FP32 Pareto landscape. A 0\% front survival rate confirms complete reorganisation of the efficient frontier under INT4, while the 21.73\% dominance flip rate indicates that one in five FP32 dominance relationships breaks down. The ground-truth KT of 0.6655 reflects moderate but meaningful ranking reordering. Despite this full structural reorganisation, the global ranking correlation is sufficiently preserved for the FP32 zero-shot surrogate to remain an effective search instrument. Finally, the Pareto rank sensitivity of 0.2404 indicates that quantization fragility is largely independent of FP32 Pareto optimality.

\begin{table}[t]
\caption{NAS-Bench-201 Performance \& Stability Metrics}
\begin{center}
\begin{tabular}{|c|c|c|c|c|}
\hline
\multicolumn{5}{|c|}{\textbf{\textit{Performance Results}}}\\
\hline
 & \multicolumn{2}{|c|}{\textbf{RS}} & \multicolumn{2}{|c|}{\textbf{MOEA}} \\
\cline{2-5} 
  & \textbf{\textit{FP32}} & \textbf{\textit{INT4}} & \textbf{\textit{FP32}} & \textbf{\textit{INT4}}\\
  & \textbf{\textit{(zero-shot)}} & & \textbf{\textit{(zero-shot)}} & \\
\hline
Acc(\%) & 0.8773 & 0.8805 & 0.8648 & 0.8749 \\
 & (0.0236)$^{\mathrm{a}}$ & (0.0248) & (0.0099) & (0.01276) \\
\hline
MFLOPs & 3.8 & 4.5 & 2.5 & 3.0 \\
 & (2.3) & (2.1) & (0.3) & (0.8) \\
\hline
Norm. Global & 0.5740 & 0.5113 & 0.6920 & 0.6481 \\
HyperVolume$^{\mathrm{b}}$ & (0.2012) & (0.1846) & (0.0238) & (0.0719) \\
\hline
\multicolumn{2}{|c|}{HyperVolume Ratio} & \multicolumn{3}{|c|}{1.12 (\textbf{RS}) / 1.07 (\textbf{MOEA})} \\
\hline
\multicolumn{2}{|c|}{Relative improvement (\%)} & \multicolumn{3}{|c|}{+12.26 (\textbf{RS}) / +6.77 (\textbf{MOEA})} \\
\hline
\multicolumn{5}{|c|}{\textbf{\textit{FP32 to INT4 Pareto Front Stability Results}}}\\
\hline
\multicolumn{2}{|c|}{Pareto Front Survival Rate (\%)} & \multicolumn{3}{|c|}{0} \\
\hline
\multicolumn{2}{|c|}{Dominance Flip Rate (\%)} & \multicolumn{3}{|c|}{21.73} \\
\hline
\multicolumn{2}{|c|}{KT-Rank Correlation} & \multicolumn{3}{|c|}{0.6655} \\
\hline
\multicolumn{2}{|c|}{Pareto Rank Sensitivity} & \multicolumn{3}{|c|}{0.2404} \\
\hline
\multicolumn{4}{l}{$^{\mathrm{a}}$Metrics format: \textit{mean\_value (std\_value)}.}\\
\multicolumn{4}{l}{$^{\mathrm{b}}$Global reference point at (acc,flops)=(0, 10M).}
\end{tabular}
\label{tab3}
\end{center}
\end{table}

\subsection{Hardware DSE Results}
In order to solidify the proposed three-stage hardware-aware NAS system, the results of the DSE process on Stage III for three Pareto front architectures are presented in Table \ref{tab4}. All three architectures converge to the same optimal CGRA4ML configuration, reflecting their shared structural profile of \textit{nor\_conv\_1x1} and \textit{skip\_connect} operations repeated across ~15 cell recurrences. The higher clock count of architecture 8592 is attributable to its single \textit{nor\_conv\_3x3} operation at the first cell node, which increases the per-cell compute workload by a factor proportional to the 3×3 vs. 1×1 kernel area ratio across all recurrences.

\begin{table}[t]
\caption{CGRA4ML DSE Results$^{\mathrm{*}}$}
\begin{center}
\begin{tabular}{|c|c|c|c|c|}
\hline
Arch. Id & PEs & $d_w$ & Clocks (cycles) & PE utilization (\%) \\
\hline
7856 & (16, 66) & 576 & 53,605 & 42.61 \\
\hline
8592 & (16, 66) & 576 & 118,690 & 45.8 \\
\hline
6854 & (16, 66) & 576 & 74,995 & 41.4 \\
\hline
\multicolumn{4}{l}{$^{\mathrm{*}}$Clocked at 250 MHz.}
\end{tabular}
\label{tab4}
\end{center}
\end{table}

\section{Conclusion}
This paper presents a three-stage hardware-aware NAS pipeline for edge AI deployment on CGRA-based accelerators, alongside an empirical study of INT4 PTQ effects on the NAS-Bench-201 Pareto space. The empirical study showed that, although INT4 PTQ causes a full Pareto structure reorganization, the dedicated INT4-trained surrogate is outperformed by its FP32 zero-shot counterpart in terms of Pareto space coverage, suggesting that the cleaner FP32 training signal transfers effectively across precision regimes. Possible future work will further optimize the hardware-agnostic frontend for more robust architecture search and extend the empirical study to other more complex mixed-precision schemes.

\vspace{12pt}
% \color{red}
% IEEE conference templates contain guidance text for composing and formatting conference papers. Please ensure that all template text is removed from your conference paper prior to submission to the conference. Failure to remove the template text from your paper may result in your paper not being published.


\begin{thebibliography}{00}
\bibitem{b1} White, Colin, et al. "Neural architecture search: Insights from 1000 papers." arXiv preprint arXiv:2301.08727 (2023).
\bibitem{b2} Li, Chaojian, et al. "Hw-nas-bench: Hardware-aware neural architecture search benchmark." arXiv preprint arXiv:2103.10584 (2021).
\bibitem{b3} Dong, Xuanyi, and Yi Yang. "Nas-bench-201: Extending the scope of reproducible neural architecture search." arXiv preprint arXiv:2001.00326 (2020).
\bibitem{b4} A. Krizhevsky and G. Hinton, “Learning multiple layers of features
from tiny images.” 2009.
\bibitem{b5} Abarajithan, G., et al. "Cgra4ml: A framework to implement modern neural networks for scientific edge computing." arXiv preprint arXiv:2408.15561 (2024).
% \bibitem{b6} Dong, Xuanyi, et al. "Nats-bench: Benchmarking nas algorithms for architecture topology and size." IEEE transactions on pattern analysis and machine intelligence 44.7 (2021): 3634-3646.
\bibitem{b7} Tu, Renbo, et al. "NAS-bench-360: Benchmarking neural architecture search on diverse tasks." Advances in Neural Information Processing Systems 35 (2022): 12380-12394.
\bibitem{b8} Benmeziane, H., et al. "A comprehensive survey on hardware-aware neural architecture search. arXiv 2021." arXiv preprint arXiv:2101.09336 (2021).
\bibitem{b9} Benmeziane, Hadjer, et al. "Multi-objective hardware-aware neural architecture search with Pareto rank-preserving surrogate models." ACM Transactions on Architecture and Code Optimization 20.2 (2023): 1-21.
\bibitem{b10} Wang, Kuan, et al. "Haq: Hardware-aware automated quantization with mixed precision." Proceedings of the IEEE/CVF conference on computer vision and pattern recognition. 2019.
\bibitem{b11} Wang, Tianzhe, et al. "Apq: Joint search for network architecture, pruning and quantization policy." Proceedings of the IEEE/CVF Conference on Computer Vision and Pattern Recognition. 2020.
\bibitem{b12} Sridhar, Sharath Nittur, et al. "SimQ-NAS: Simultaneous Quantization Policy and Neural Architecture Search." arXiv preprint arXiv:2312.13301 (2023).
\bibitem{b13} Gao, Tianxiao, et al. "QuantNAS: Quantization-aware neural architecture search for efficient deployment on mobile device." Proceedings of the IEEE/CVF Conference on Computer Vision and Pattern Recognition. 2024.
\bibitem{b14} Lin, Yujun, Mengtian Yang, and Song Han. "NAAS: Neural accelerator architecture search." arXiv preprint arXiv:2105.13258 (2021).
\bibitem{b15} Mylonas, Eleftherios, et al. "A Unified FPGA/CGRA Acceleration Pipeline for Time-Critical Edge AI: Case Study on Autoencoder-Based Anomaly Detection in Smart Grids." Electronics 15.2 (2026): 414.
\end{thebibliography}
\end{document}